\documentclass{rescience}
\usepackage{subfigure}
\begin{document}

\begin{tikzpicture}[remember picture, overlay]
  \node [shape=rectangle, draw=darkred, fill=darkred, yshift=-34mm,
        anchor=north west, minimum width=3.75cm, minimum height=10mm]
        at (current page.north west) {};
  \node [text=white, anchor=center, yshift=-39mm, xshift=1.875cm]
         at (current page.north west) {\small RESCIENCE C};
\end{tikzpicture}

{\let\newpage\relax\maketitle} \maketitle

\marginnote{
  \footnotesize \sffamily
  \textbf{Edited by}\\
  \ifdefempty{\editorNAME}{\textcolor{darkgray}{(Editor)}}
             {\editorNAME\ifdefempty{\editorORCID}{}{$^{\orcid{\editorORCID}}$}}\\
  ~\\
  \ifdefempty{\reviewerINAME}{}
  {
  \textbf{Reviewed by}\\
  \ifdefempty{\reviewerINAME}{\textcolor{darkgray}{}}
             {\reviewerINAME\ifdefempty{\reviewerIORCID}{}{$^{\orcid{\reviewerIORCID}}$}\\}
  \ifdefempty{\reviewerIINAME}{\textcolor{darkgray}{}}
             {\reviewerIINAME\ifdefempty{\reviewerIIORCID}{}{$^{\orcid{\reviewerIIORCID}}$}\\}
  ~\\
  }
  \textbf{Received}\\
  \ifdefempty{\dateRECEIVED}{---}{\dateRECEIVED}\\
  ~\\
  \textbf{Published}\\
  \ifdefempty{\datePUBLISHED}{---}{\datePUBLISHED}\\
  ~\\
  \textbf{DOI}\\
  \ifdefempty{\articleDOI}{---}{\articleDOI}
}

\newcommand{\container}[1]{\def\@container{#1}}
\begin{container}
  \afterpage {
    \begin{statement}
      \scriptsize \sffamily
      \hrule \vskip .5em
      Copyright © \articleYEAR~\authorsABBRV,
      released under a Creative Commons Attribution 4.0 International license.
      
      Correspondence should be addressed to
      \contactNAME~(\href{mailto:\contactEMAIL}{\contactEMAIL})
  
      The authors have declared that no competing interests exist.

      \ifdefempty{\codeURL}{}
      {Code is available at
      \href{\codeURL}{\detokenize\expandafter{\codeURL}}\ifdefempty{\codeDOI}{.}{ -- DOI \doi{\codeDOI}.}\ifdefempty{\codeSWH}{.}{ -- SWH  \href{https://archive.softwareheritage.org/\codeSWH/}{\detokenize\expandafter{\codeSWH}}.}}

      \ifdefempty{\dataURL}{}
      {Data is available at
      \href{\dataURL}{\detokenize\expandafter{\dataURL}}\ifdefempty{\dataDOI}{.}{ -- DOI \doi{\dataDOI}.}}

      \ifdefempty{\reviewURL}{}
     {Open peer review is available at \href{\reviewURL}{\detokenize\expandafter{\reviewURL}}.}
    \end{statement}
  }
\end{container}



\begin{abstract}\label{abst}
    \noindent
    Large vision-language models (VLMs) are shown to learn rich joint image-text representations enabling high performances in relevant downstream tasks. However, they fail to showcase their quantitative understanding of objects, and they lack good counting-aware representation.  This paper conducts a reproducibility study of `Teaching CLIP to Count to Ten' \citep{paiss2023teaching}, which presents a method to finetune a CLIP model \citep{CLIP} to improve zero-shot counting accuracy in an image while maintaining the performance for zero-shot classification by introducing a counting-contrastive loss term. We contribute to the existing methods by improving the model's performance on a smaller subset of their training data with lower computational resources. We verify these claims by reproducing their study with our own open-source code. The implementation can be found at \url{https://github.com/SforAiDl/CountCLIP}. 
    
    \end{abstract}
    \section{Introduction}\label{intro}
    In recent years, the development of large Vision-Language Models (VLMs) has significantly propelled the field of representation learning in computer vision, with models like CLIP \citep{CLIP} and BASIC \citep{BASIC} showcasing their ability to learn robust joint image-text representations. Tasks such as zero-shot classification, segmentation, image captioning, and text-to-image generation benefit from the compositional understanding capabilities of such models. However, these models struggle on counting tasks such as matching an image representation based on the number of the specified objects present, to the text representation of the count corresponding to the image's caption. The popularity of text-to-image and text-to-video models such as Stable Diffusion Video \citep{blattmann2023stable}, Sora \citep{videoworldsimulators2024}, Lumiere \citep{bartal2024lumiere}, has risen and such models rely heavily on CLIP for their image-text representations. Count-aware models like \citep{paiss2023teaching} shall greatly enhance the abilities of these existing architectures, enabling models to produce highly accurate videos with the correct number of entities.
    
    \citet{paiss2023teaching} suggested shifting the training objective to discriminate between the correct and the incorrect captions associated with the object counts given an image by introducing a counting loss term $L_{count}$ to the VLM's loss function. The paper reported a pipeline to create training data consisting of counting images. A \textit{counting image} is an image whose caption accurately reflects the number of entities in the image, as shown in Figure \ref{fig:train}(b). This data is then used to fine-tune a pre-trained VLM by contrasting the representation of the correct caption with that of the synthetically generated counterfactual caption, where the count in the correct caption is randomly swapped to an incorrect one. This way, the VLM is trained to align images with its true count captions and discriminate against the incorrect ones. The trained model is evaluated on the \textit{CountBench} benchmark, an object counting benchmark introduced by the paper. Noting its importance to the contributions to VLM's, this paper seeks to reproduce and build upon their work. 
    
    This paper primarily aims at reproducibility while also advancing the existing research. Our keys contributions include:
    \begin{enumerate}
        \item We improved the model's performance on a smaller subset, 640 times smaller than that of the paper's training data and with lower computational resources, beating the baseline by 1.38\%.
        \item We created and released our own counting image training dataset, making it publicly accessible. 
        \item We created and made public a more comprehensive version of the CountBench benchmark by manually including images from non-functional URLs.
    \end{enumerate}
    \section{Materials and Methods}\label{methods}
    
    \subsection{Dataset}\label{method:data}
    The LAION-400M \citep{LAION} dataset is used to obtain the counting image dataset. Of the 400 million images, they obtained roughly 200,000 counting images. Due to our size and computational constraints, we passed over 2 million images and obtained $\sim$2,000 counting images. We have made our code and counting set public. As described in the paper, we looked for the sentences that contained the numbers \textit{"two"} to \textit{"ten"} in words. Images were passed through the YOLOv8 \citep{YOLOv8} object detector, and the count of the most frequent entity was checked with the count in the caption. If they are equal, the image-text pair is added to the counting set, as seen in Figure \ref{fig:train}(a) .
    
     The method to generate the counting set was highly unfeasible when running over the full dataset. The training data could have been made public with the URLs being public. The large number of non-functioning URLs in the original dataset was a major bottleneck. At least 30 out of 540 images were unavailable in the \textit{CountBench} due to the URLs of the images being defunct. We contacted the authors of the papers, and they stated that while they possess the image files, it was against their company policy to share raw images sourced from publicly available data. The benchmark is the only one of its kind, and it was carefully curated, to ensure class balance. As a result, any future work measuring counting accuracy must use this benchmark in addition to other methods for fairer and more reproducible comparisons.

    \subsection{Training the model}\label{method:train}
    \begin{figure}
        \centering
        \subfigure[]{\includegraphics[width=0.7\textwidth]{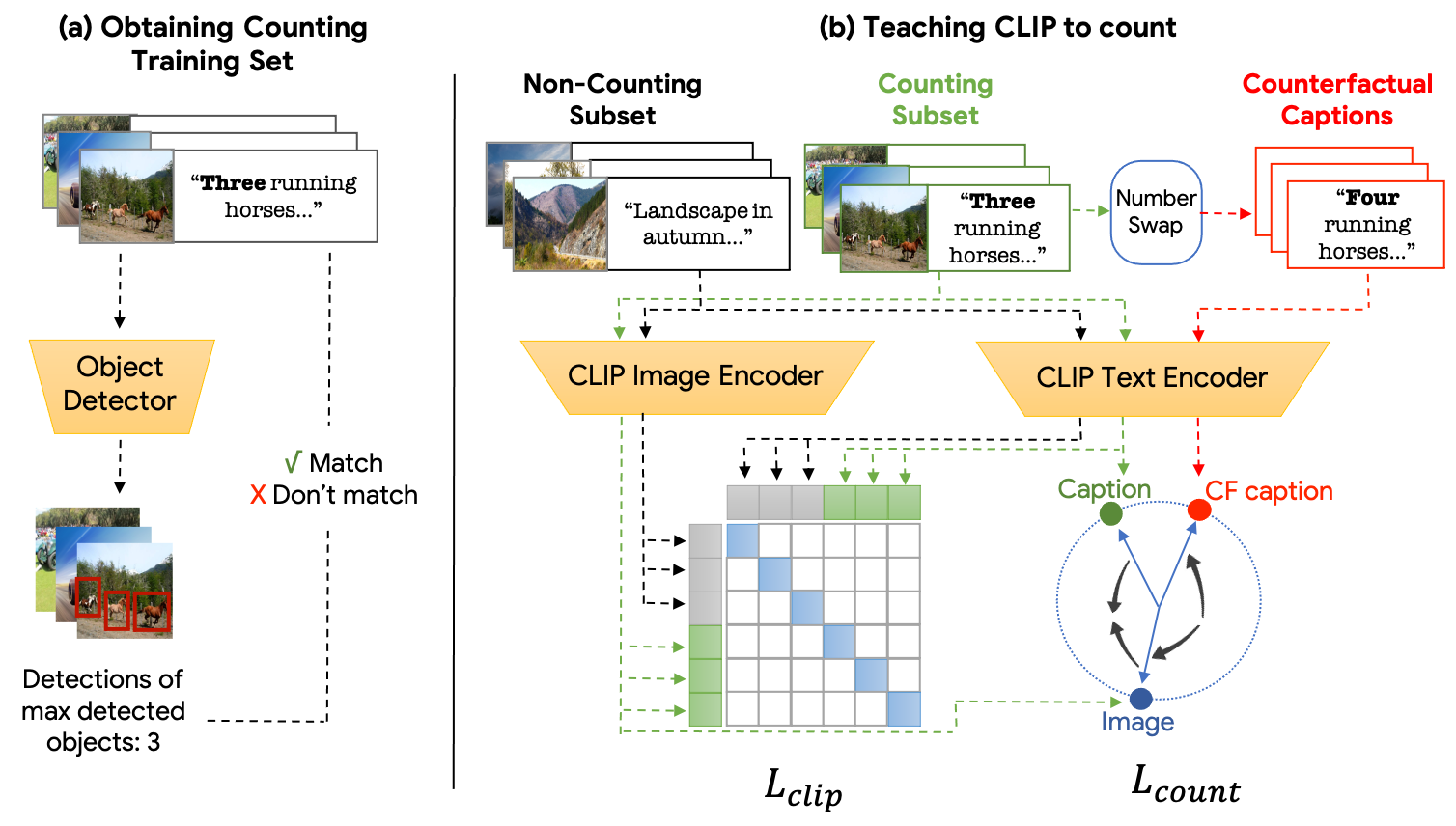}} 
        \subfigure[]{\includegraphics[width=0.34\textwidth]{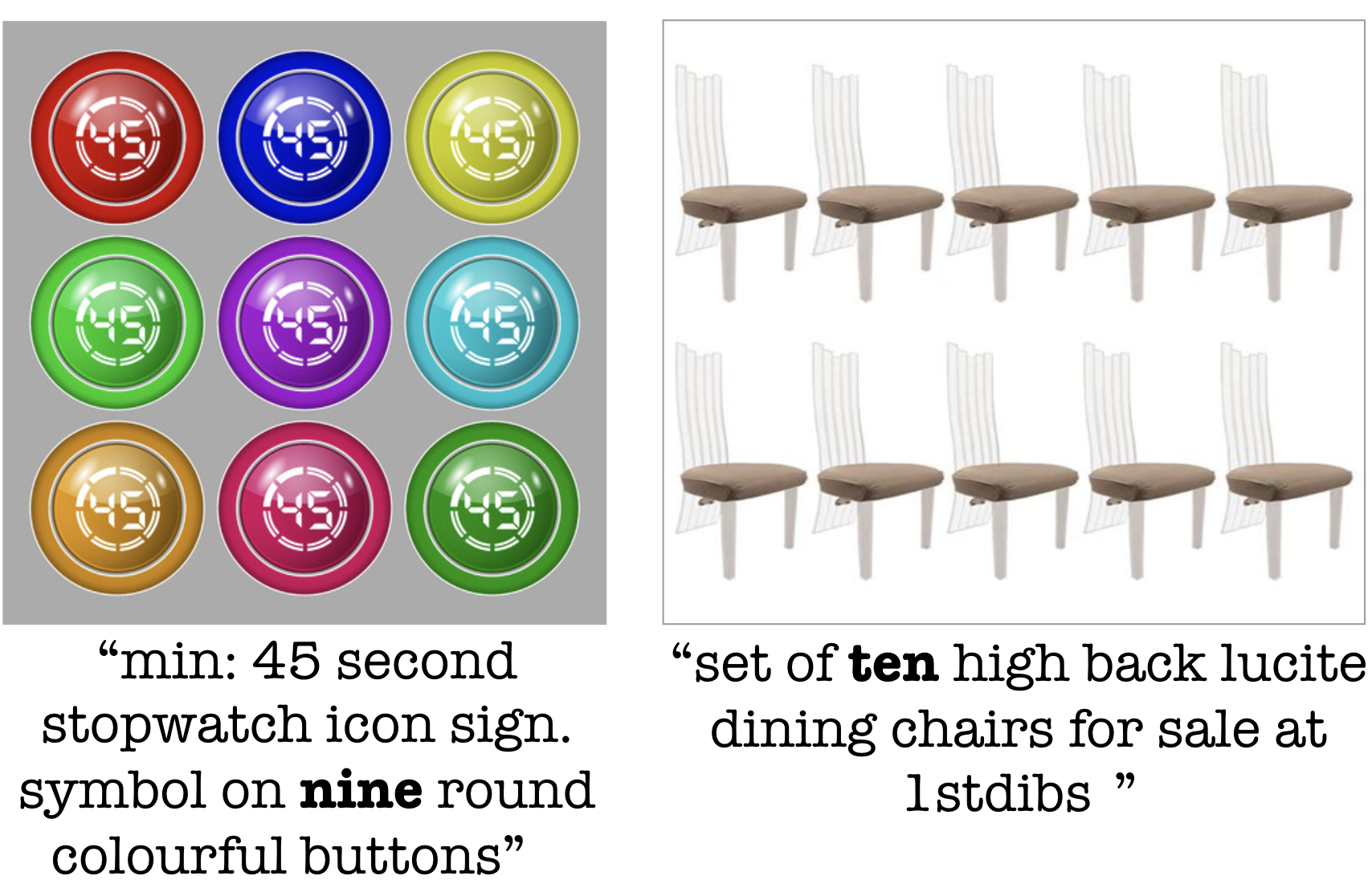}} 
        \subfigure[]{\includegraphics[width=0.34\textwidth]{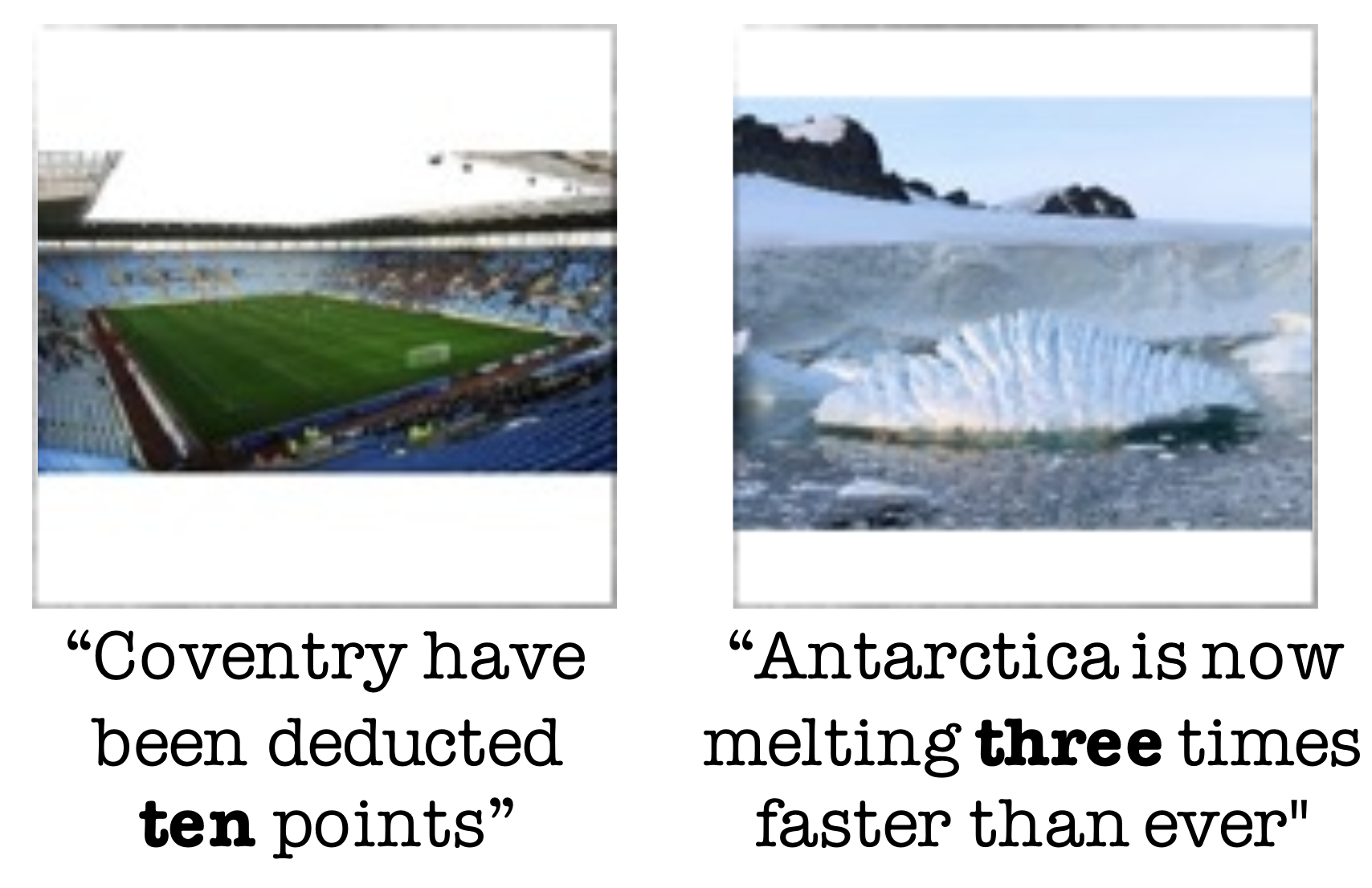}} 
        \caption{Figure copyright \citet{paiss2023teaching}.
                    (a) Training setup
                    (b) Examples of counting images
                    (c) Examples of noncounting images 
                }
    
    \label{fig:train}
    \end{figure}
    
    The authors in the original paper utilized a dual loss function, combining the regular contrastive loss of CLIP $(L_{CLIP})$ with a counting-designated loss $(L_{count})$, weighted by the hyperparameter $\lambda$ (Equation \ref{eq:1}).
    
    \begin{equation}
        \label{eq:1}
        L = L_{CLIP} + \lambda L_{count}
    \end{equation}
    
    The fine-tuning process involved training the model on two different training sets: (i) LAION-400M \citep{LAION}, an extensive dataset collected from the web comprising general images and captions, and (ii) a filtered numbered training set $C$, as described in Section \ref{method:data}, containing samples with object counts spelled out in the captions. While the regular contrastive loss $L_{CLIP}$ was calculated on all samples, the counting loss $L_{count}$ was specifically computed for samples from set $C$, with $N$ counting image-text pairs in a batch. For each image-text pair $(i_{k}, t_{k})$ in set $C$, a counterfactual caption $t^{CF}_{k}$ was automatically generated by replacing the number in the original caption $t_{k}$ with a different random number (e.g., for an image consisting of four parrots, the corresponding true caption
    “four parrots” can be counterfactualized with “seven parrots”). 
    
    During each step of the training process, the tuples $(i_{k}, t_{k}, t^{CF}_{k})^{N}_{k=1}$ are fed into CLIP's text and image encoders, resulting in the generation of their respective embeddings  $(ei_{k}, et_{k}, et^{CF}_{k})^{N}_{k=1}$. The contrastive loss $L_{count}$ was then computed to ensure a high similarity score between the image and the original caption and a low similarity score with the counterfactual caption (Equation \ref{eq:2}). This loss function encouraged the model to learn the relationship between the spelled-out number in the caption and the number of objects it referred to. This is shown in Figure \ref{fig:train}. 
    
    \begin{equation}
        \label{eq:2}
        L_{count} = - \frac{1}{N} \sum_{k=1}^{N} \log\frac{\exp(ei_{k}\cdot et_{k})}{\exp(ei_{k}\cdot et_{k}) + \exp(ei_{k}\cdot et^{CF}_{k})}
    \end{equation}
    
    Furthermore, negative samples were exclusively used in the counting objective $L_{count}$, rather than being added to the batch for the existing contrastive loss $L_{CLIP}$. This selective approach was adopted to better weigh the impact of negative samples on the counting objective. To reproduce the experimental setup accurately, we followed the described procedures, including creating counterfactual captions and the training regime on the specified datasets.

    \subsection{Balancing lambda}\label{exp:bal}
    In section \ref{method:train}, the hyperparameter $\lambda$ was introduced to adjust the weightage of $L_{CLIP}$ and $L_{count}$ in the final loss function. In the paper, the optimal lambda was reported to be 1.We introduce a new scheme for setting the hyperparameter $\lambda$ by balancing it on the frequencies of the classes. The main motivation behind this is the severe imbalance in the counting training data for both our counting set as well as \citet{paiss2023teaching}'s. The paper used a form of undersampling, as they had very a large amount of data avaliable to them. This scheme of choosing $\lambda$ ensures that more focus is given to the less frequent class. This technique allows us to use this modified loss on smaller datasets like ours. 
    
    We present three ways to balance lambda, $\lambda_{norm}$,  $\lambda_{modal}$, $\lambda_{log}$. The $\lambda_{norm}$ in equation (\ref{eq:3}) setting the focus in proportion to the frequency of classes. The lambda is normalised as it is always a fraction of $\lambda_{0}$.
    
    \begin{equation}
        \label{eq:3}
        \lambda_{norm}(class) = (1-\frac{n_{class}}{n_{total}})\lambda_{0}
    \end{equation}
    
    The $\lambda_{modal}$ (given by Equation (\ref{eq:4})) sets more focus on less frequent classes, while ensuring that the most frequent (modal) class has the minimum value $\lambda_{0}$.
    \begin{equation}
        \label{eq:4}
        \lambda_{modal}(class) = \frac{n_{modal}}{n_{class}}\lambda_{0}
    \end{equation}
    
\begin{figure}
        \centering
        \subfigure[]{\includegraphics[width=0.3\textwidth]{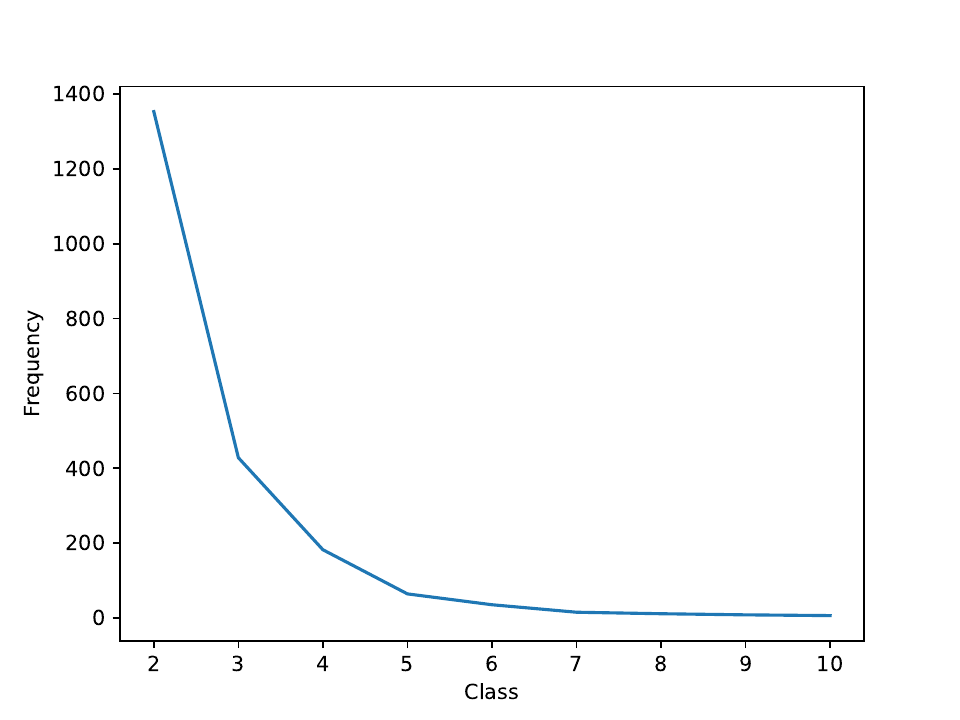}} 
        \subfigure[]{\includegraphics[width=0.3\textwidth]{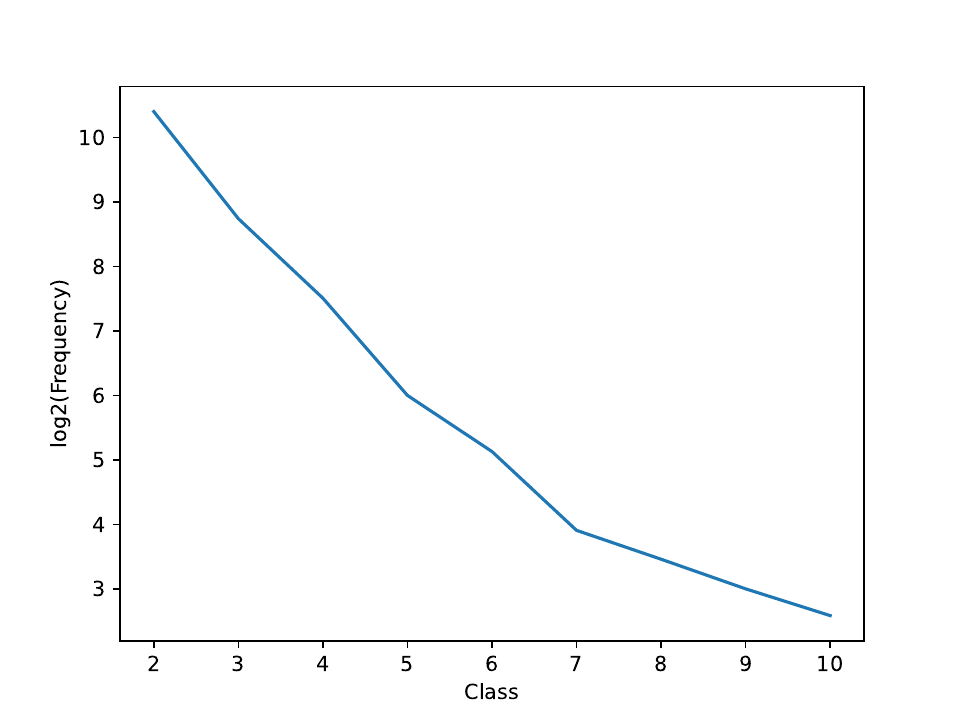}}  
        \subfigure[]{\includegraphics[width=0.3\textwidth]{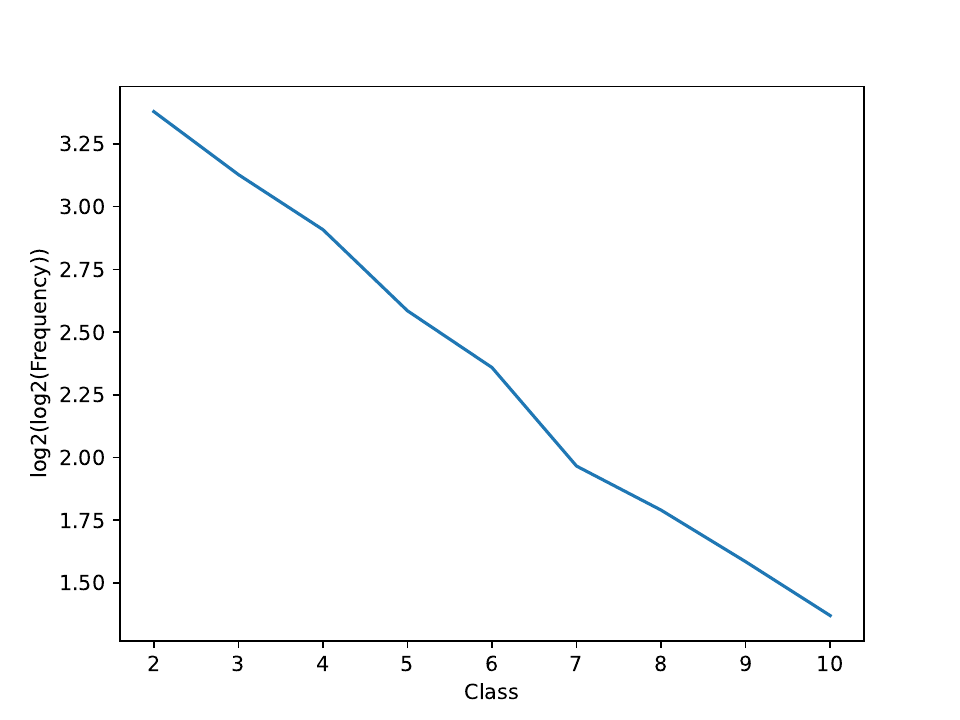}}

        \caption{
                    (a) Class frequency
                    (b) $log_{2}(frequency)$
                    (c) $log_{2}(log_{2}(frequency))$ 
                }
        \label{fig:llog}
    \end{figure}

    The $\lambda_{log}$ in Equation (\ref{eq:5}) and $\sigma$ in Equation (\ref{eq:6}) are computed for all classes. $\sigma_{min},\sigma_{max}$ are the minimum and maximum values of $\sigma$ after computing it for each class respectively.
    The training dataset had a class imbalance resembling an exponential curve. By taking the logarithmic transformation as in Equation (\ref{eq:6}), we can apply a linear scaling to the $\lambda$, while avoiding overfitting on the least frequent classes.
    The class distributions are plotted in figure \ref{fig:llog}(a), and resembles an exponential curve.
    On taking the logarithm, we see that the distribution is now close to linear. We apply the logarithm again to make the distribution as linear as possible for our training data. This is seen in the figure \ref{fig:llog}.
    We have tested the model by taking the logarithm once (i.e., $log_2(\frac{n_{total}}{n_{class}})$) for computing $\sigma$, however, it yielded poorer results, so we proceeded with the transformation as per Equation \ref{eq:6}.

    \begin{equation}
        \label{eq:5}
        \lambda_{log}(class) = (\frac{\sigma_{class}-\sigma_{min}}{\sigma_{max}}+1)\lambda_{0}
    \end{equation}
    
    \begin{equation}
        \label{eq:6}
        \sigma(class) = log_2(log_2(\frac{n_{total}}{n_{class}}))
    \end{equation}

    $n_{class}$ is the number of examples of the given class, and $n_{total}$ is the total number of counting images in the training dataset. The base lambda $\lambda_{0}$ is set to 1. During train time, we pass the counting caption to a function to get its appropriate value of $\lambda_{balanced}$ based on its count.
    
    \subsection{CountPlus}\label{exp:countplus}
    The loss function introduced in the paper (Equation \ref{eq:2}) contrasted the correct caption with the counterfactual caption, where the count in the correct caption is randomly swapped to an incorrect one. We experiment by changing the loss so that it contrasts the correct caption with all possible counterfactual captions, where the number is swapped with all possible incorrect values.  
    
    \begin{equation}
        \label{eq:7}
        L_{count+} = - \frac{1}{N} \sum_{k=1}^{N} \log\frac{\exp(ei_{k}\cdot et_{k})}{\exp(ei_{k}\cdot et_{k}) + \sum_{j=2;j\neq count}^{10}\exp(ei_{k}\cdot et^{CF}_{k})}
    \end{equation}

    \section{Evaluation}\label{exp}

    The paper also published a new image-text counting benchmark \textit{CountBench} for evaluating a model’s understanding of object counting, which we used to benchmark our model. The data is carefully curated and class-balanced, with 540 images. We create embeddings of all the possible combinations for the captions of the images, take the embeddings of the 9 captions and images, and compute the similarity score with each caption. The caption with the highest similarity score is the predicted class. We have normalised the embeddings before the dot product in all the results reported in Table \ref{tab:widgets}. Figure \ref{fig:confmatrix1} and Figure \ref{fig:confmatrix2} gives the confusion matrices for the tested configurations.

    \newpage
    
    \section{Results}\label{res}
    
    \begin{figure}
    \subfigure[]{\includegraphics[width=0.325\textwidth]{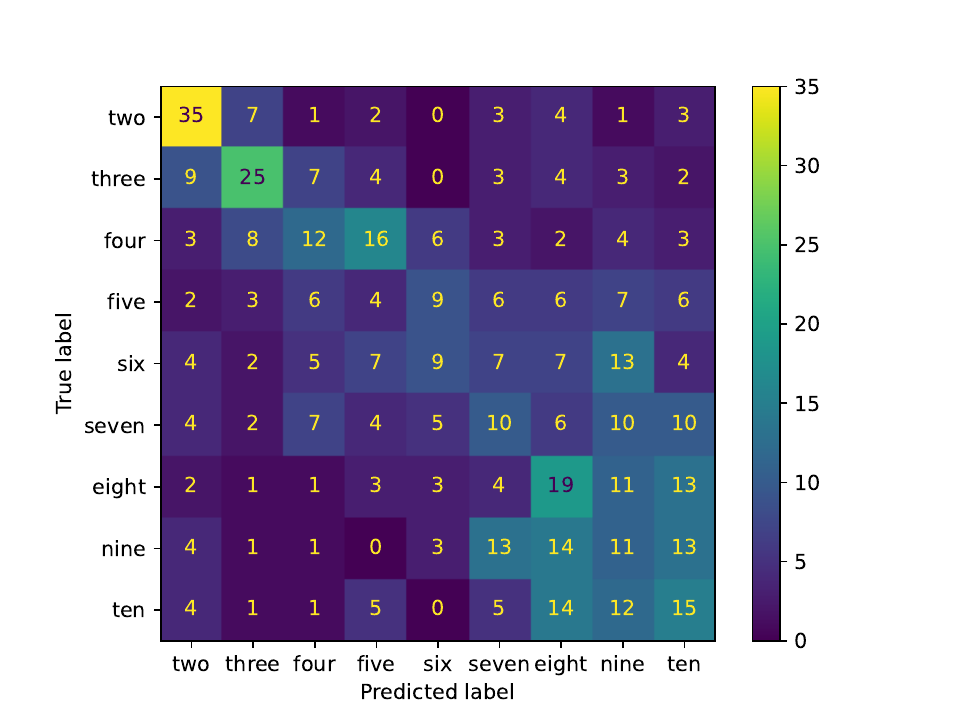}} 
    \subfigure[]{\includegraphics[width=0.325\textwidth]{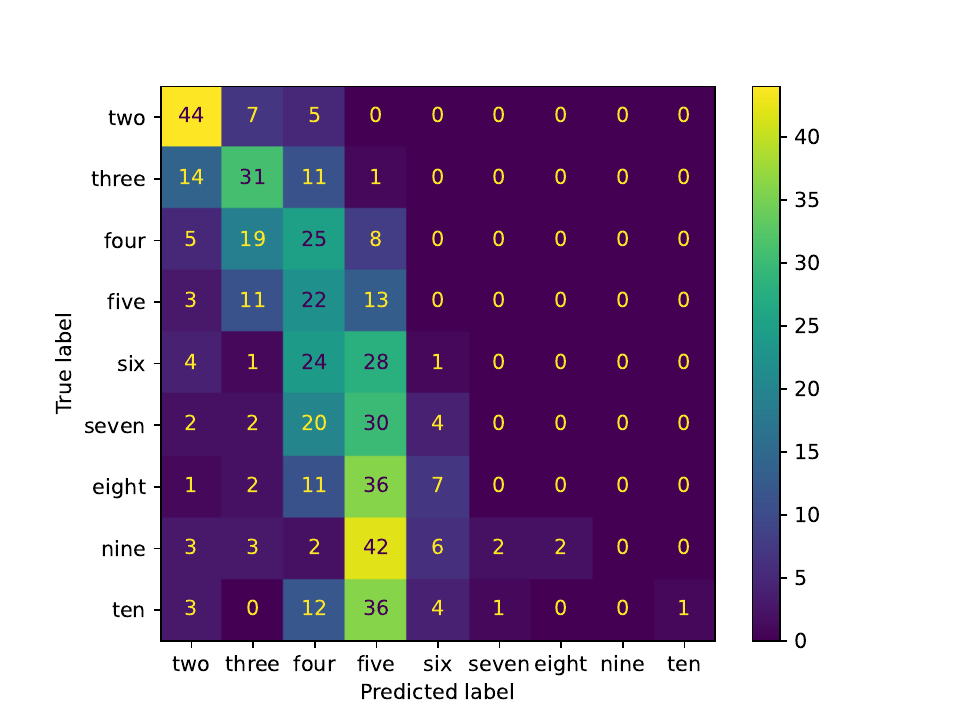}} 
    \subfigure[]{\includegraphics[width=0.325\textwidth]{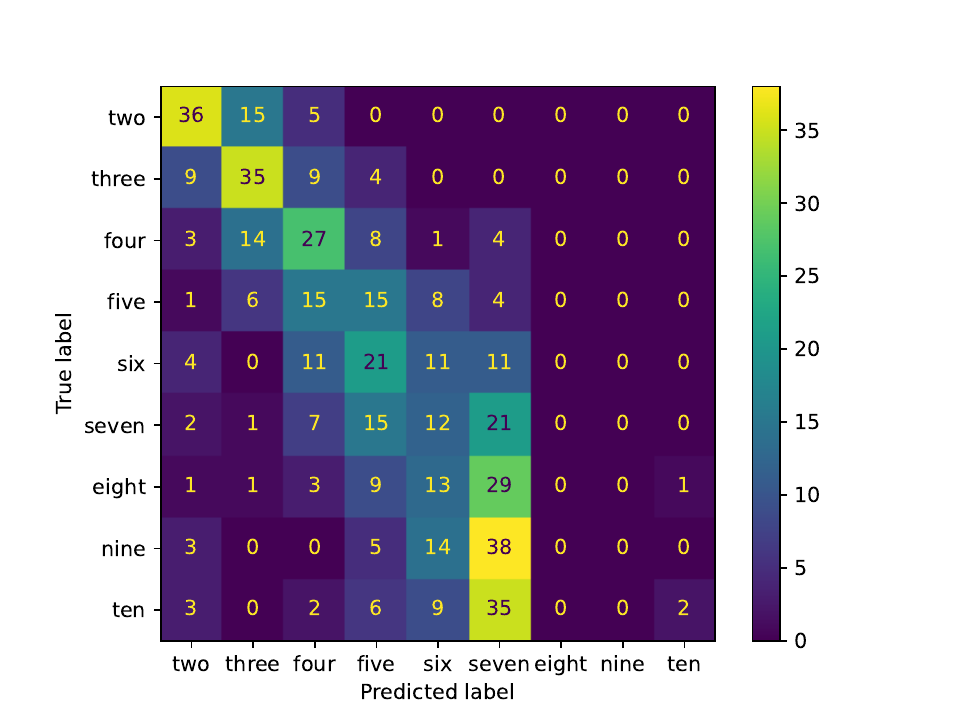}}
    \centering
    \subfigure[]{\includegraphics[width=0.325\textwidth]{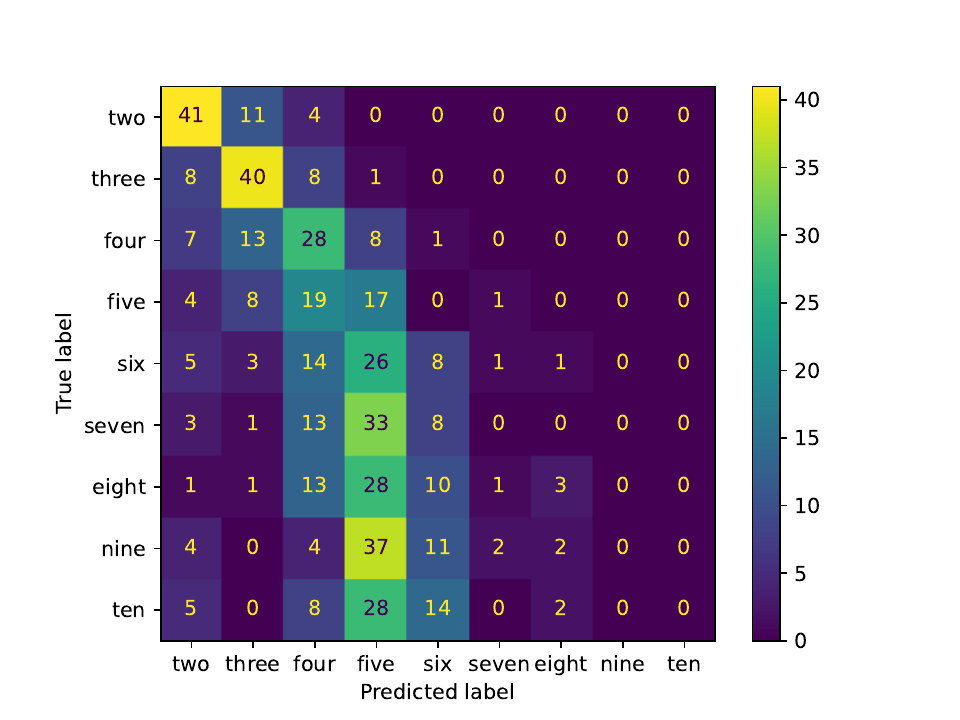}} 
    \subfigure[]{\includegraphics[width=0.325\textwidth]{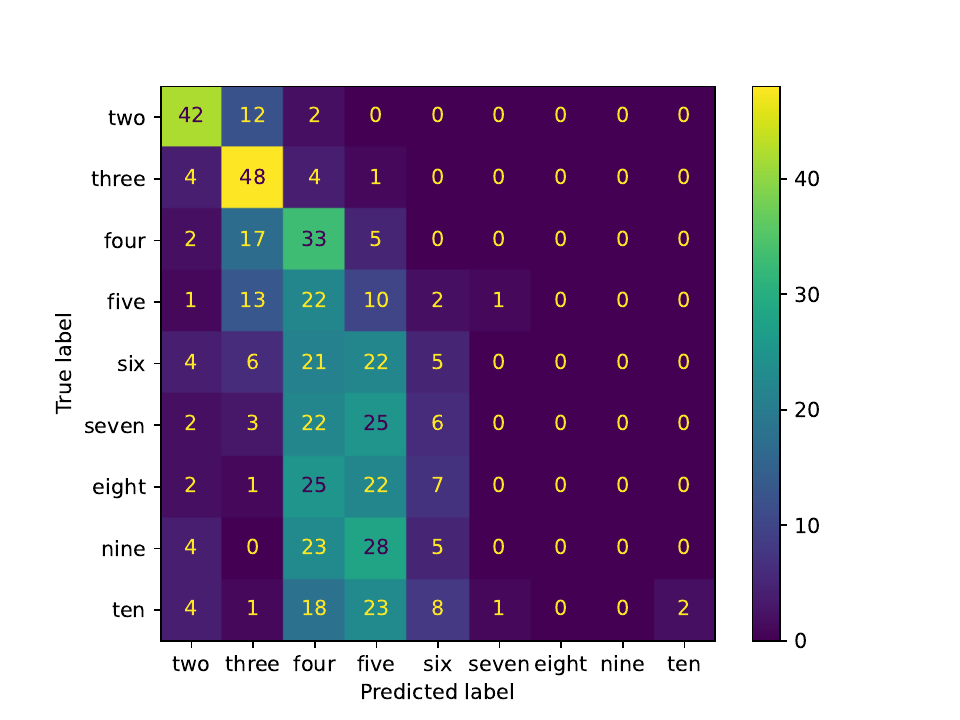}} 
    
    \caption{Confusion matrices with early stopping for: (a)baseline
                    (b) scheduler and $\lambda = 1 $ (base model) 
                    (c) scheduler and $\lambda_{modal}$ and $L_{count+}$
                    (d) scheduler and $\lambda_{norm}$ and $L_{count+}$ 
                    (e) scheduler and $\lambda_{log}$ and $L_{count+}$}
    \label{fig:results}
    
    \end{figure}
    
    The baseline accuracy for a CLIP B/32 model on the CountBench dataset is 27.5\% with no training. The confusion matrices for the main comparisons are given in Figure \ref{fig:results}. Please refer to Figure \ref{fig:confmatrix1} and Figure \ref{fig:confmatrix2} of Appendix \ref{appendix} for the confusion matrices for the remaining experiments. The accuracies as a percentage are reported in Table \ref{tab:widgets}.  \citet{paiss2023teaching} reported the results on both CLIP and BASIC. However, in this paper, we only focus on training CLIP. We have used a learning rate of $5\mathrm{e}{-6}$, and 20,000 steps (10 epochs), and a linear warmup in the first half of the steps with a cosine scheduler in the other half. The difference is in the amount of data used to train the model, $b_{size}$ is the batch size, $p$ is the proportion of counting images in a batch, and $n_{count}$ is the total number of counting images in the training data. \citet{paiss2023teaching}: $b_{size} = 32,768 ,p = 1/32, n_{count} = 200,000$; Ours: $b_{size} = 5, p = 1/5, n_{count} = 2000$.

    \begin{table}[h] 
    \centering
    \begin{tabular}{l|c|c}
    Configuration & Acc. (end of $10^{th}$ epoch)& Max Acc. (early stop)\\\hline
    $\lambda = 1 $ with no scheduler & 21.81 & 25.15\\
    scheduler and $\lambda = 1 $ (base model) & 22.59 & 25.93\\
    $\lambda_{norm}$ & 21.61 & 25.74\\
    scheduler and $\lambda_{norm}$ & 25.54 & 25.54\\ 
    scheduler and $\lambda_{norm}$ and $L_{count+}$ & \textbf{26.92} & 26.92\\ 
    $\lambda_{modal}$ & 20.83 & 26.13\\
    scheduler and $\lambda_{modal}$ & 19.84 & 26.33\\
    scheduler and $\lambda_{modal}$ and $L_{count+}$& 26.33 & \textbf{28.88}\\
    $\lambda_{log}$ & 19.25 & 24.56\\
    scheduler and $\lambda_{log}$ & 21.81 & 27.90\\
    scheduler and $\lambda_{log}$ and $L_{count+}$& 22.59 & 27.50\\
    
    \end{tabular}
    \caption{Zero-shot counting accuracy for the configurations for a CLIP-B/32 model}
    \label{tab:widgets}
    \end{table}
    
    \newpage
    \section{Discussion}\label{discuss}
    
    We found adapting the auxiliary loss weight as a class weight with an appropriate scheme is an effective way to boost the performance of the model in scenarios with extreme class imbalance and low training dataset sizes.
    We have also found that changing the counting objective to a multiclass classification loss leads to better performance when used with a scheme for balancing lambda, even exceeding or matching the baseline by upto 1.38\% despite the severe lack of data.  
    We performed an ablation of the learning rate scheduler and observed a gain in both maximum accuracy (using early stopping) and the end of 10th epoch in most cases.
    \newline
    \newline
    Better schemes can be devised to improve accuracy, a limitation of using such schemes is that, while it may increase the overall accuracy, it may also weaken the accuracy of classes that are more data rich. The confusion matrices in Figure \ref{fig:results} show that the models rarely predict the higher-numbered classes, which may be due to the models not learning to classify those classes due to a lack of training data available for those classes (refer to Figure \ref{fig:llog}). Specifically, classes 7-10 each had training data less 20. Our models have learned to classify the counts much better than the baseline for classes with more data (i.e., classes 2-6). With more diverse training data, our methods are more likely to get better results on all the classes.

    \section{Conclusions}\label{conclusion}
    We carried out a reproducibility study of the paper `Teaching CLIP to Count to Ten' (\citealt{paiss2023teaching}), that introduced a new loss function to fine-tune VLM's to make them adept at counting tasks and released a benchmark for the same. In our efforts to reproduce the paper, we modified the loss function for a small training dataset ($\sim640$ times smaller than the dataset used in the paper). Despite the small amount of data our modifications have been shown to improve the performance of the model, surpassing the baseline. We also thoroughly audited the benchmark CountBench published in the paper, finding that $\sim30$ of the 540 images in the dataset are missing. We have made our datasets and code public, in an effort to make the work on count-aware VLMs to be more accessible to the research community. 
    \newpage
    \appendix
    \section{Visualising Results}
    \label{appendix}
    
\begin{figure}
        \centering
        \subfigure[]{\includegraphics[width=0.3\textwidth]{conf_matrices/baseline.pdf}} 
        \subfigure[]{\includegraphics[width=0.3\textwidth]{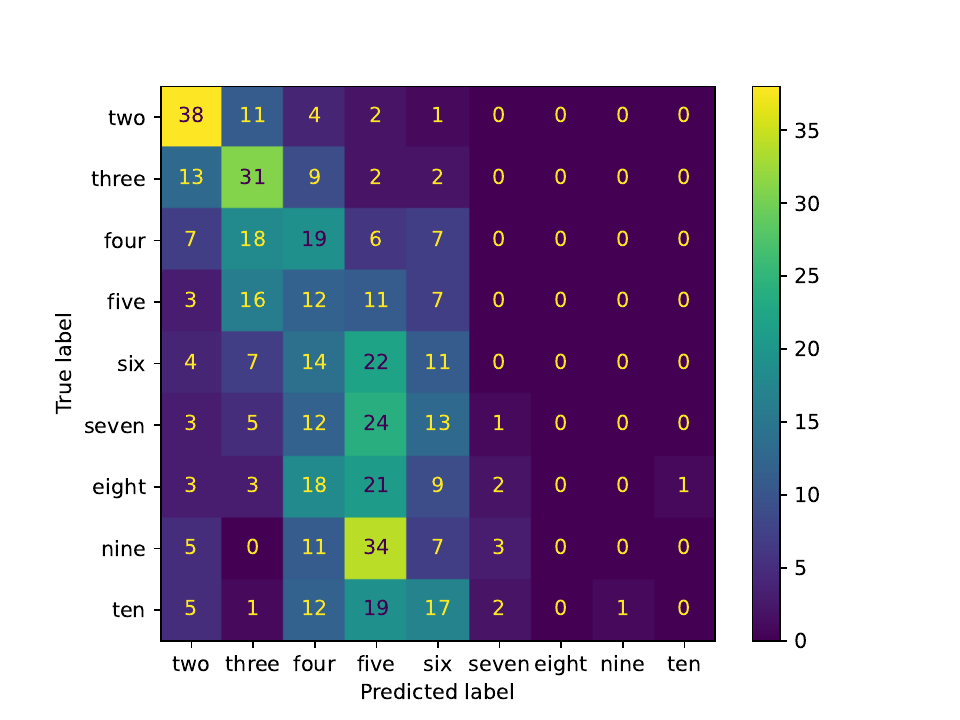}}  
        \subfigure[]{\includegraphics[width=0.3\textwidth]{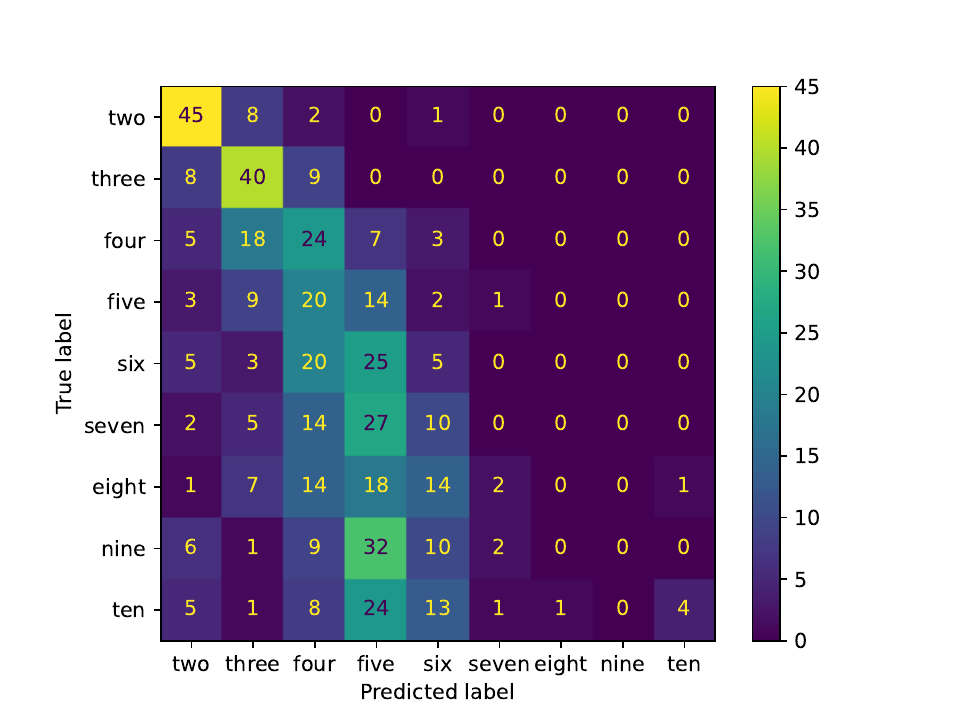}}
        
        \subfigure[]{\includegraphics[width=0.3\textwidth]{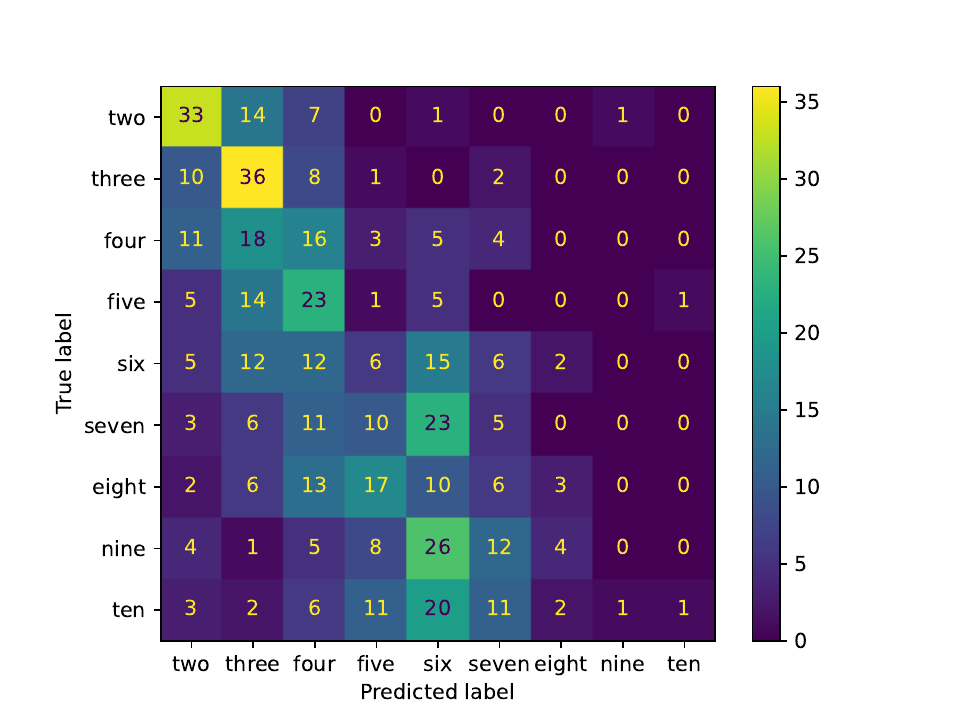}}
        \subfigure[]{\includegraphics[width=0.3\textwidth]{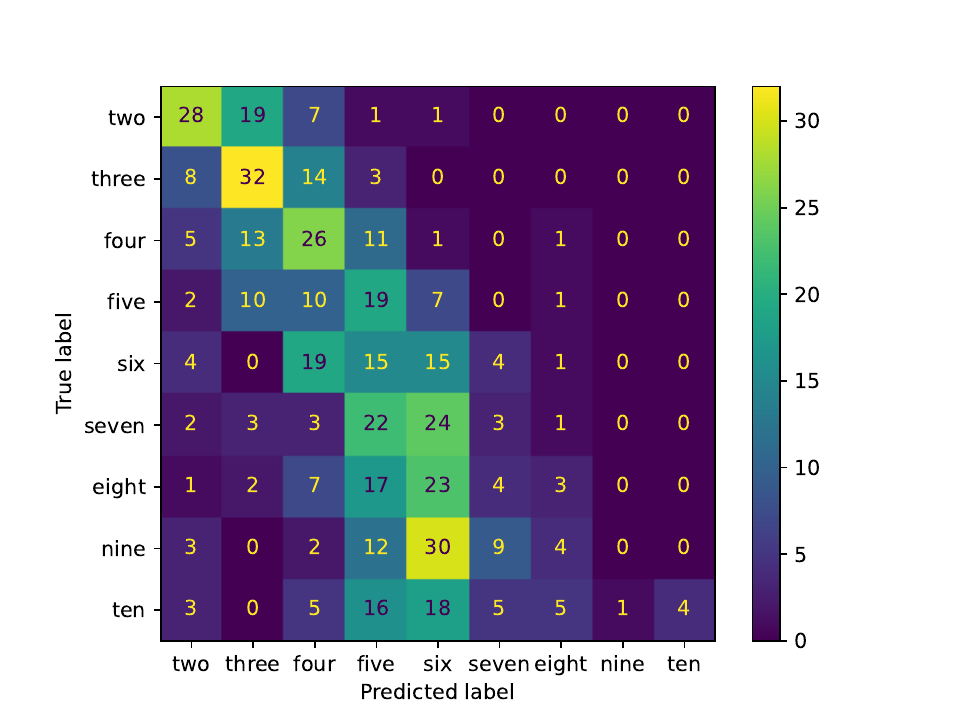}} 
        \subfigure[]{\includegraphics[width=0.3\textwidth]{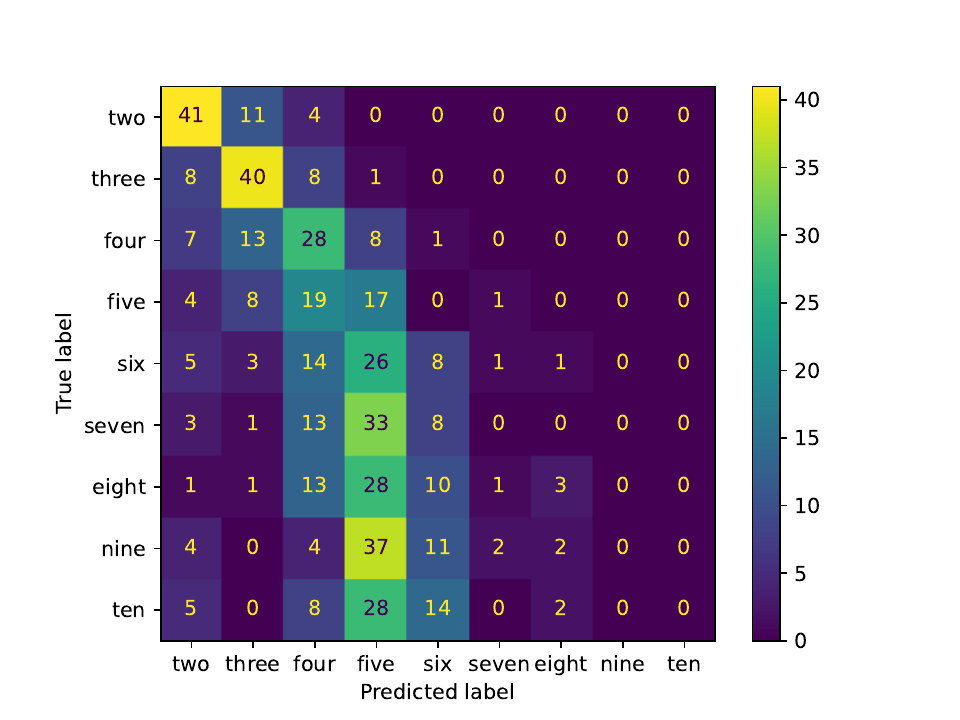}}
        
        \subfigure[]{\includegraphics[width=0.3\textwidth]{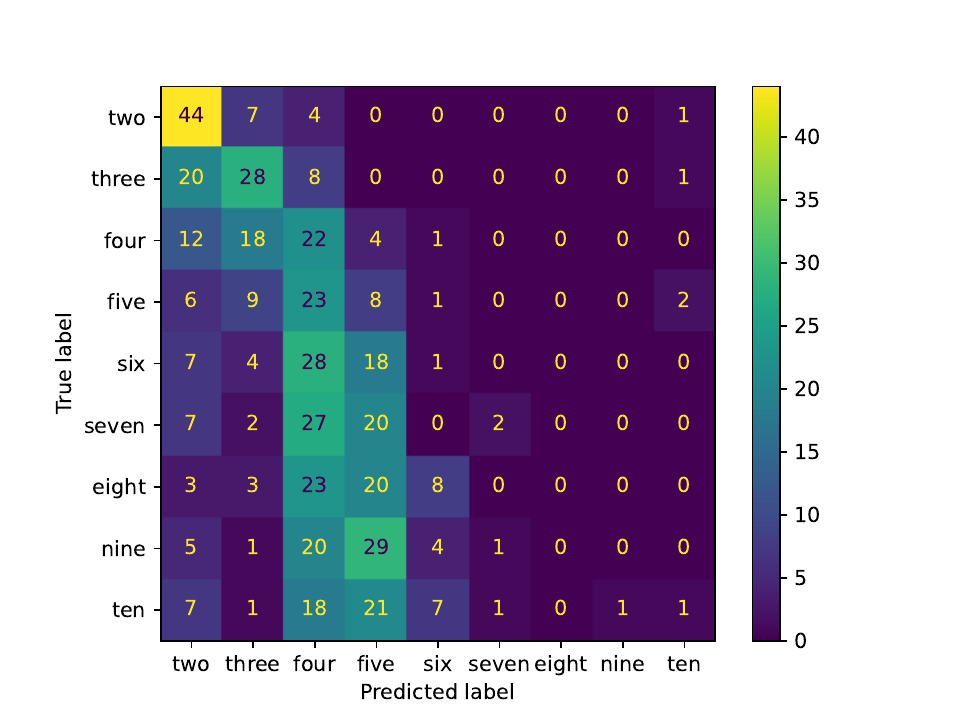}}
        \subfigure[]{\includegraphics[width=0.3\textwidth]{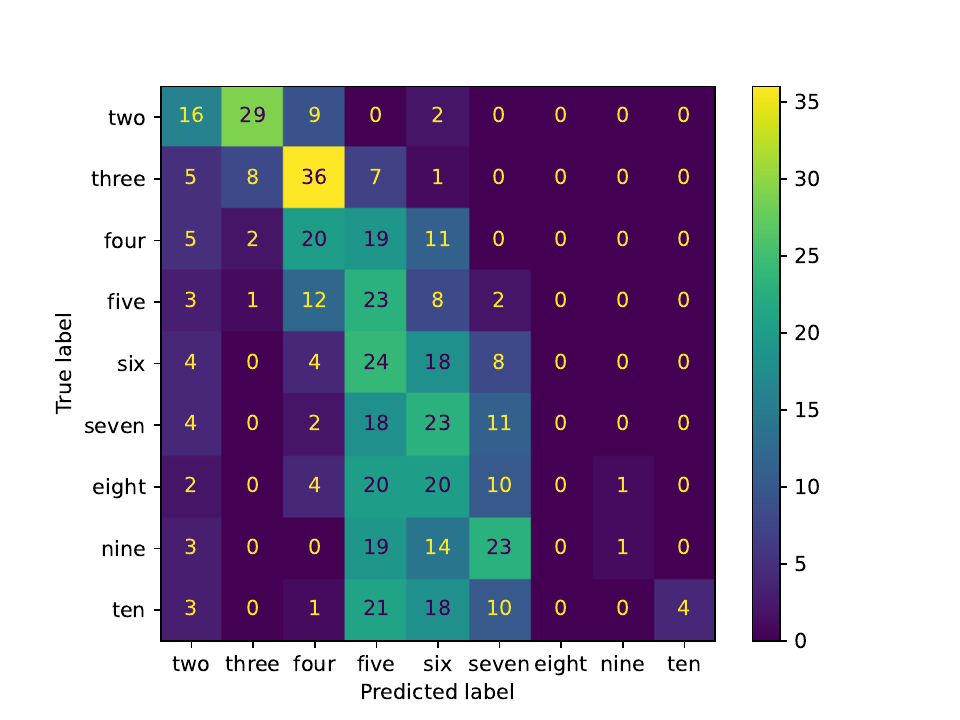}}
        \subfigure[]{\includegraphics[width=0.3\textwidth]{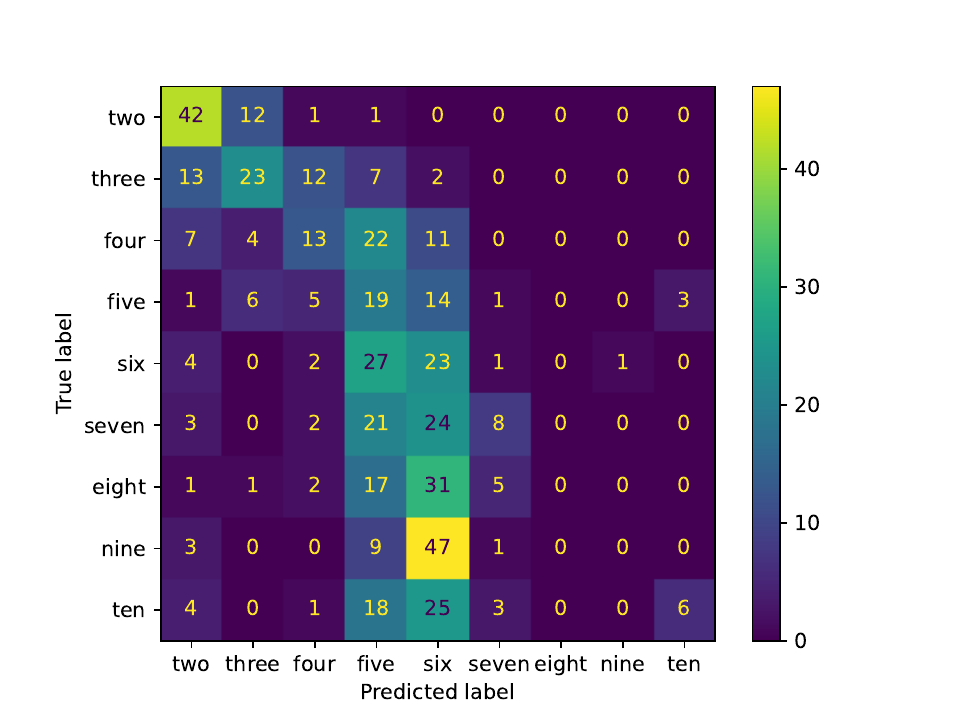}} 
        
        \subfigure[]{\includegraphics[width=0.3\textwidth]{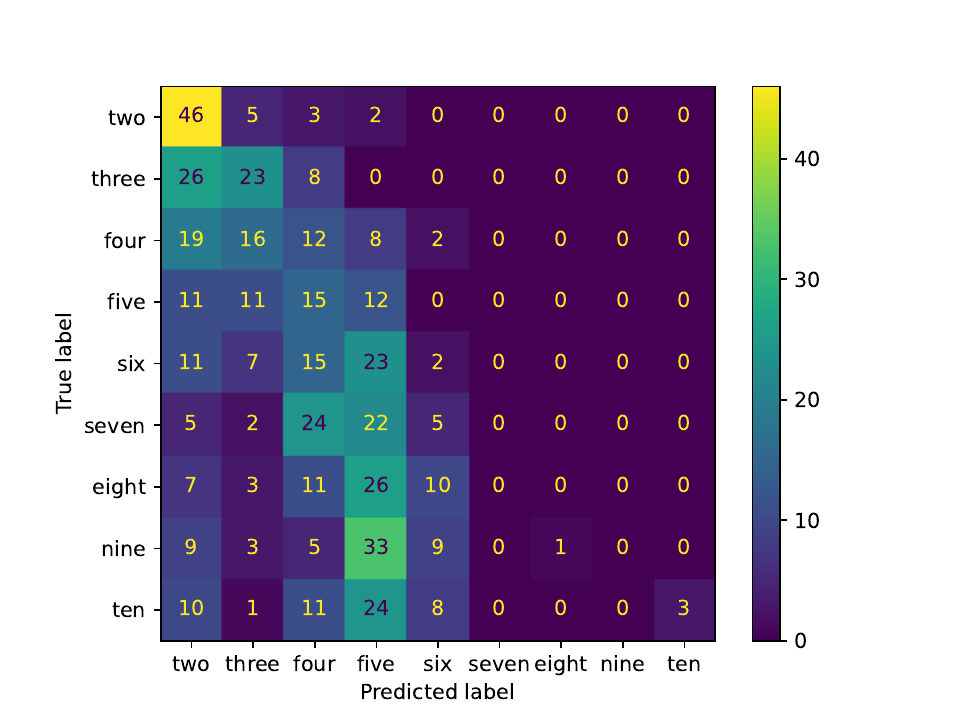}} 
        \subfigure[]{\includegraphics[width=0.3\textwidth]{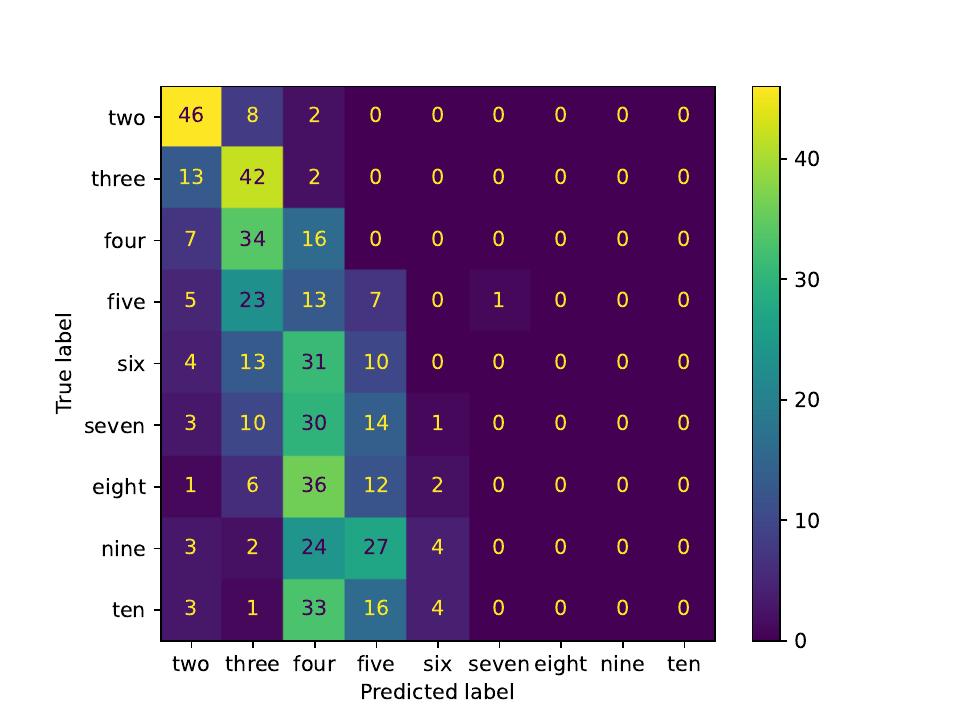}}
        \subfigure[]{\includegraphics[width=0.3\textwidth]{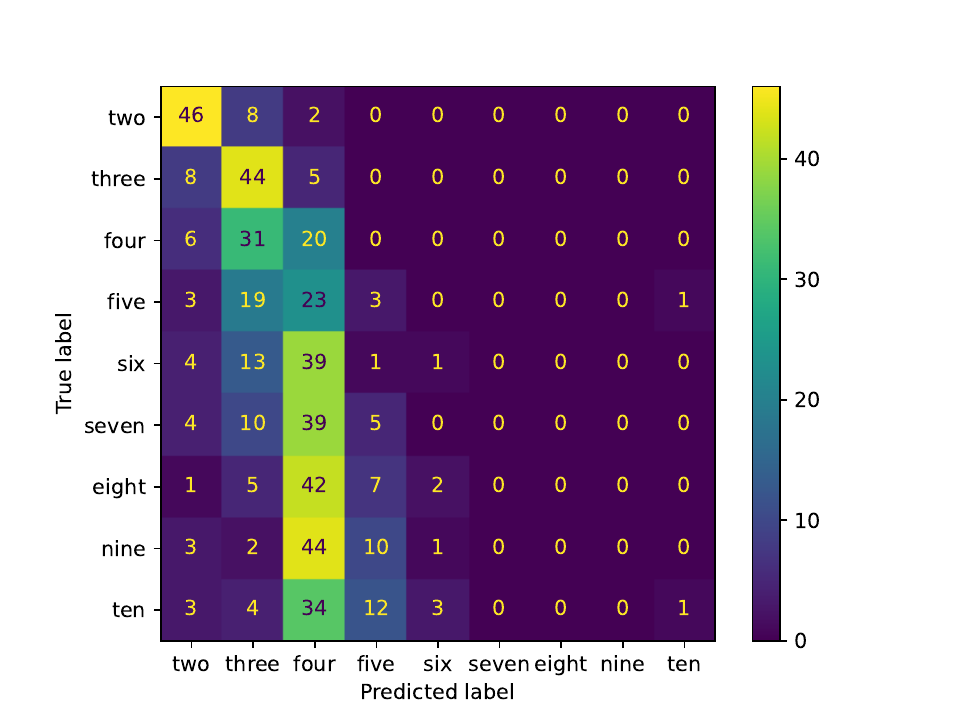}}

         \caption{Confusion matrices for models trained till the end of the $10^{th}$ epoch for: (a)baseline
                    (b) $\lambda = 1 $ with no scheduler 
                    (c) scheduler and $\lambda = 1 $ (base model) 
                    (d) $\lambda_{norm}$ 
                    (e) scheduler and $\lambda_{norm}$ 
                    (f) scheduler and $\lambda_{norm}$ and $L_{count+}$ 
                    (g) $\lambda_{modal}$ 
                    (h) scheduler and $\lambda_{modal}$
                    (i) scheduler and $\lambda_{modal}$ and $L_{count+}$
                    (j) $\lambda_{log}$ 
                    (k) scheduler and $\lambda_{log}$
                    (l) scheduler and $\lambda_{log}$ and $L_{count+}$}
        \label{fig:confmatrix1}
        
    \end{figure}

\begin{figure}
        \centering
        \subfigure[]{\includegraphics[width=0.3\textwidth]{conf_matrices/baseline.pdf}} 
        \subfigure[]{\includegraphics[width=0.3\textwidth]{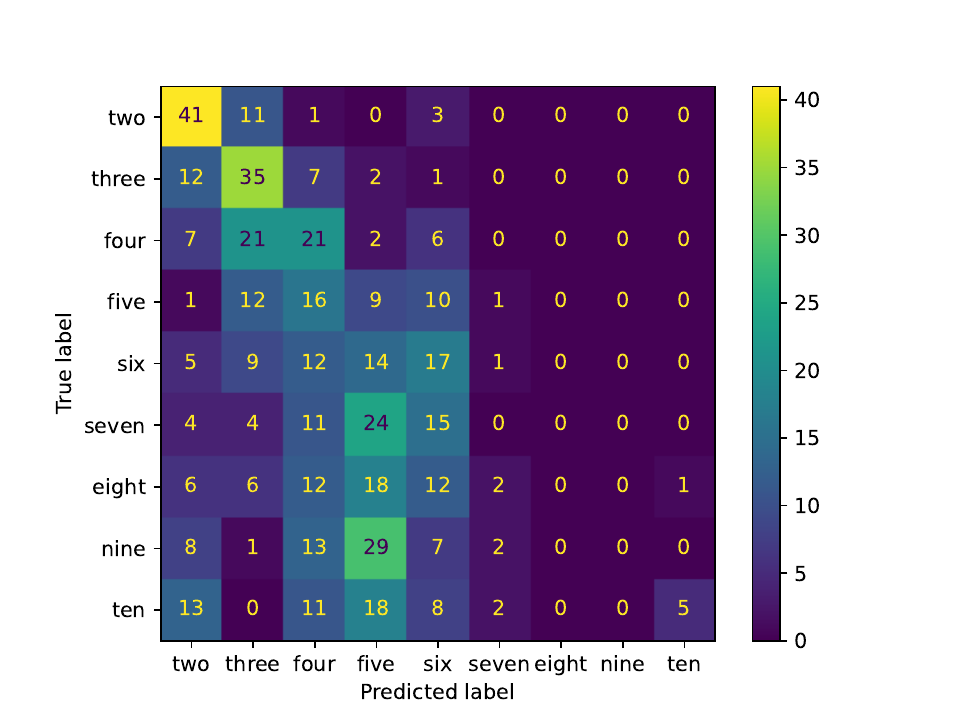}}  
        \subfigure[]{\includegraphics[width=0.3\textwidth]{conf_matrices/max/norm_sch.pdf}}
        
        \subfigure[]{\includegraphics[width=0.3\textwidth]{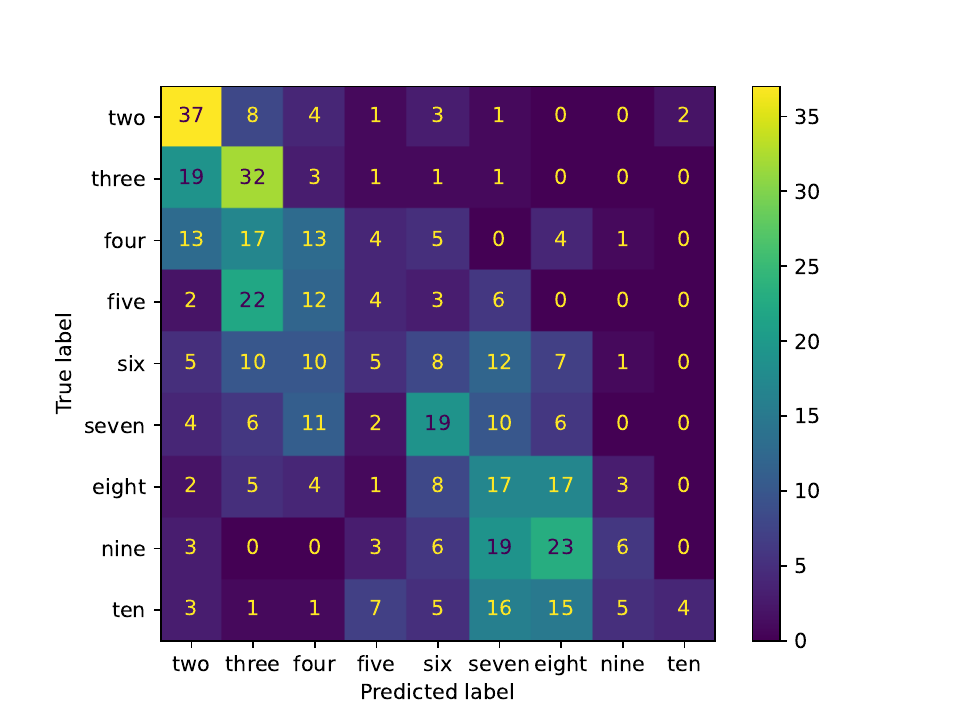}}
        \subfigure[]{\includegraphics[width=0.3\textwidth]{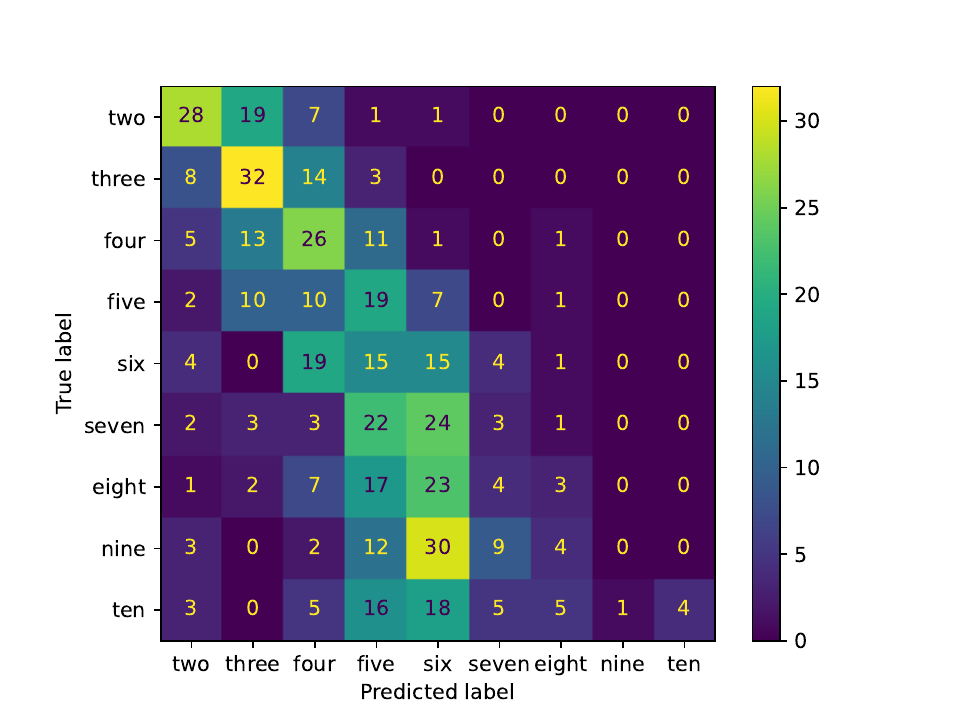}} 
        \subfigure[]{\includegraphics[width=0.3\textwidth]{conf_matrices/max/norm_sch_Lnorm_cplus.pdf}}
        
        \subfigure[]{\includegraphics[width=0.3\textwidth]{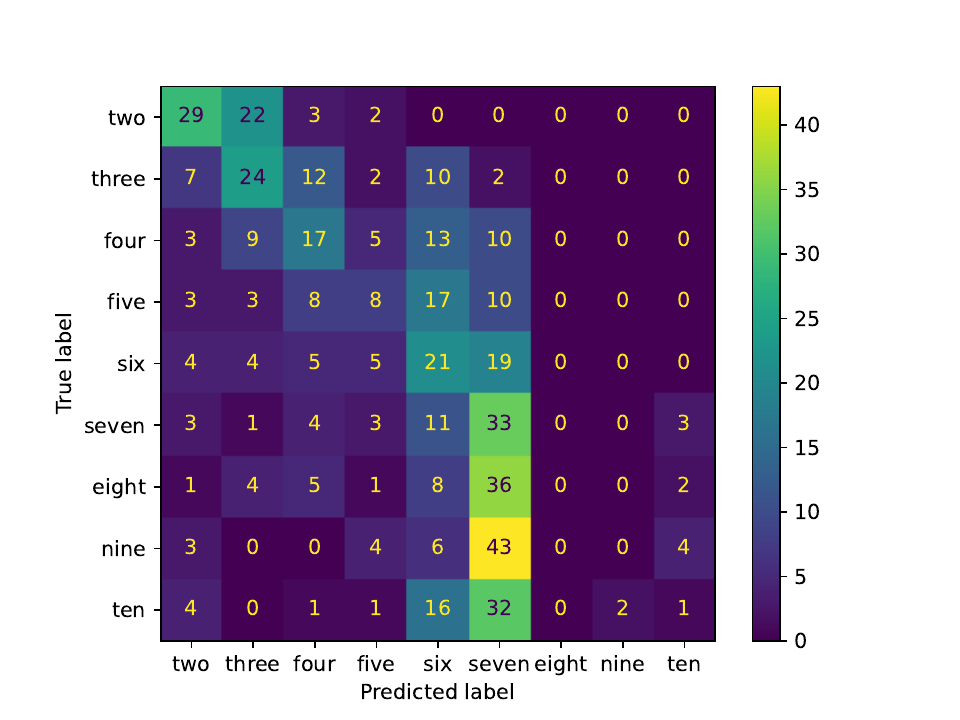}}
        \subfigure[]{\includegraphics[width=0.3\textwidth]{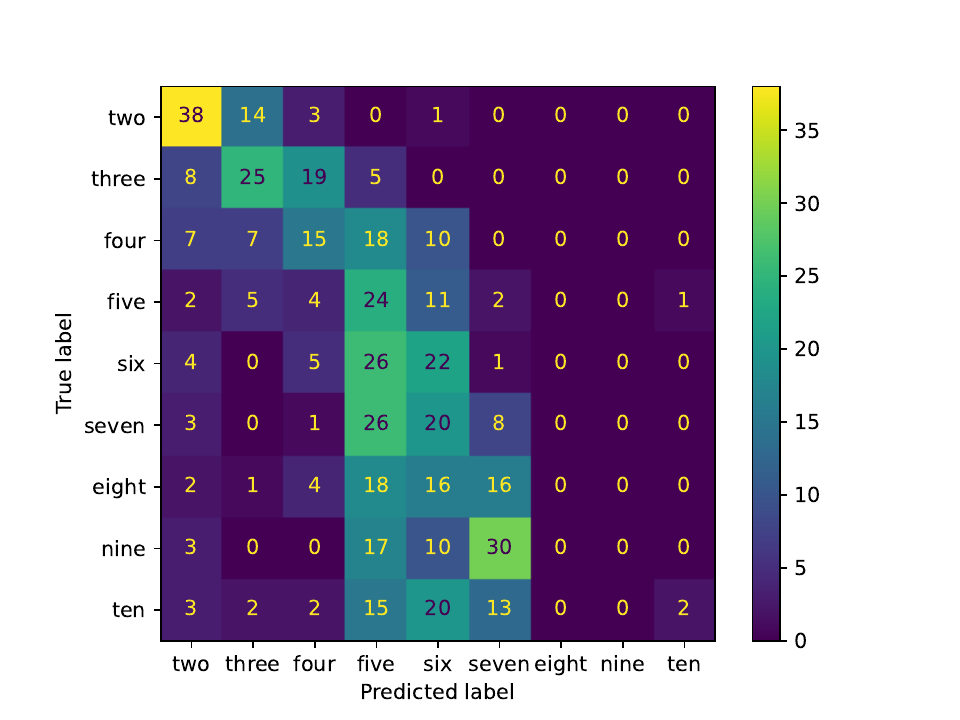}}
        \subfigure[]{\includegraphics[width=0.3\textwidth]{conf_matrices/max/norm_sch_Lmodal_cplus.pdf}} 
        
        \subfigure[]{\includegraphics[width=0.3\textwidth]{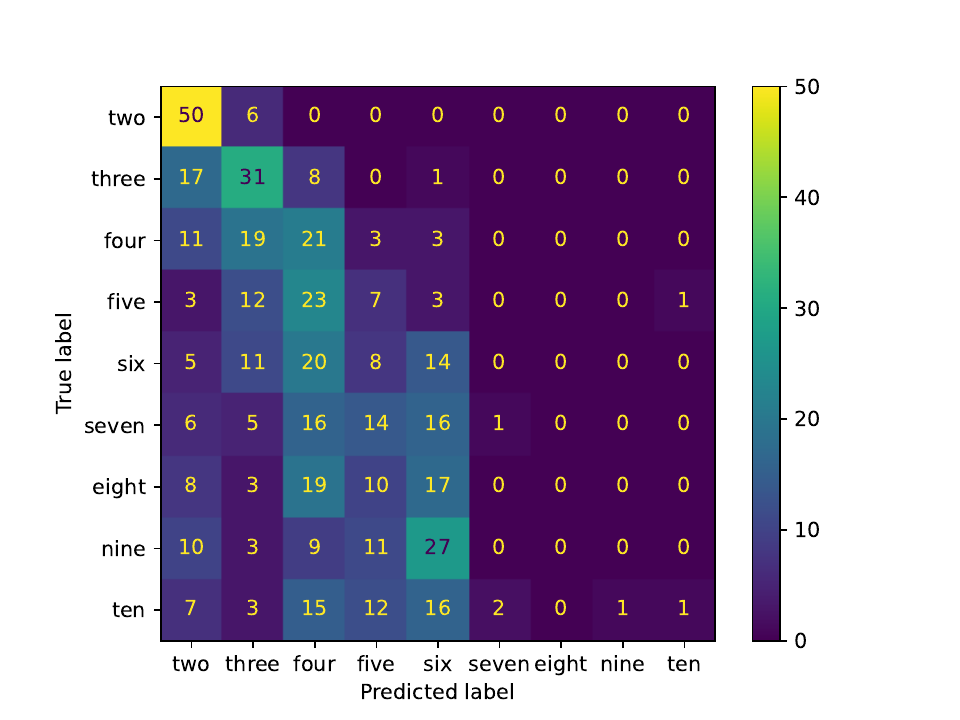}} 
        \subfigure[]{\includegraphics[width=0.3\textwidth]{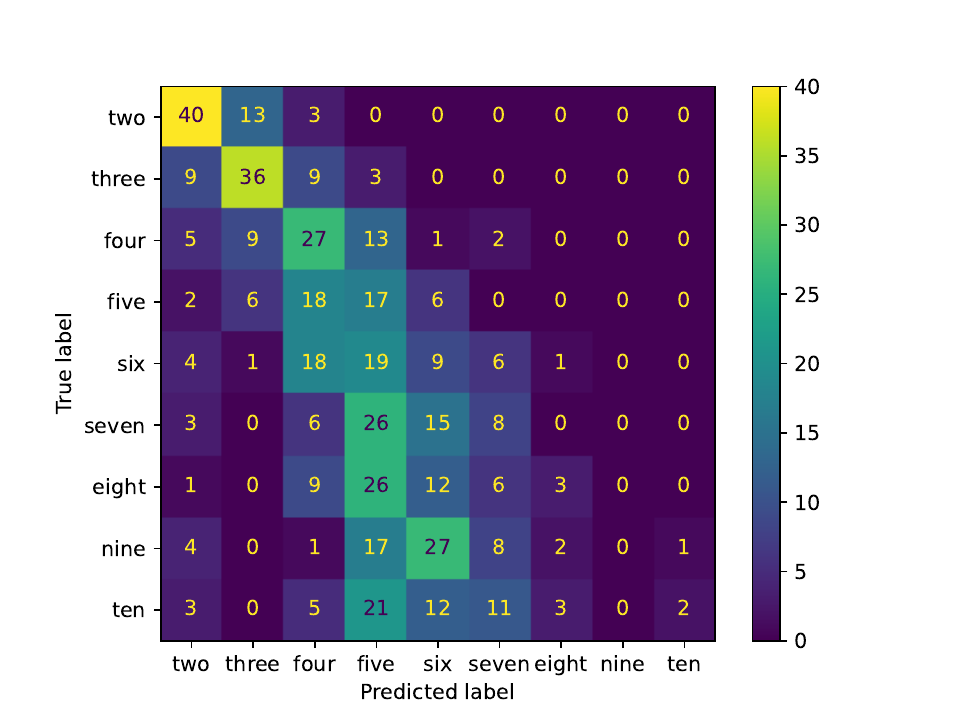}}
        \subfigure[]{\includegraphics[width=0.3\textwidth]{conf_matrices/max/norm_sch_Llog_cplus.pdf}}

        \caption{Confusion matrices for an early stopping mechanism selecting models with maximum validation accuracy for :(a)baseline
                    (b) $\lambda = 1 $ with no scheduler 
                    (c) scheduler and $\lambda = 1 $ (base model) 
                    (d) $\lambda_{norm}$ 
                    (e) scheduler and $\lambda_{norm}$ 
                    (f) scheduler and $\lambda_{norm}$ and $L_{count+}$ 
                    (g) $\lambda_{modal}$ 
                    (h) scheduler and $\lambda_{modal}$
                    (i) scheduler and $\lambda_{modal}$ and $L_{count+}$
                    (j) $\lambda_{log}$ 
                    (k) scheduler and $\lambda_{log}$
                    (l) scheduler and $\lambda_{log}$ and $L_{count+}$}
        \label{fig:confmatrix2}
    \end{figure}

    This section the contains the confusion matrices for all the experiments as mentioned in Table \ref{tab:widgets} in Section 4. Figure \ref{fig:confmatrix1} contains the confusion matrices for models till the end of the 10th epoch. Figure \ref{fig:confmatrix2} contains the confusion matrices for an early stopping mechanism selecting models with maximum validation accuracy.
    
\hypersetup{linkcolor=black,urlcolor=darkgray}
\renewcommand\emph[1]{{\bfseries #1}}
\setlength\bibitemsep{0pt}
\printbibliography

\end{document}